\documentclass[conference]{IEEEtran}
\IEEEoverridecommandlockouts
\usepackage{cite}
\usepackage{graphicx}
\usepackage{textcomp}

\usepackage{amsmath,amssymb,amsfonts}

\usepackage{array}
\usepackage{booktabs}
\usepackage{makecell}
\usepackage{multirow}
\usepackage{colortbl}
\usepackage{xcolor}
\definecolor{headerbg}{gray}{0.93}

\usepackage{algorithm}
\usepackage{algorithmic}

\usepackage{stfloats} % allows top-of-page 2-column floats
\usepackage{url}

\usepackage{etoolbox}
\AtBeginEnvironment{equation}{\small}
\AtBeginEnvironment{align}{\small}
\AtBeginEnvironment{gather}{\small}
\AtBeginEnvironment{multline}{\small}

\makeatletter
\patchcmd{\subsection}{\@startsection{subsection}{2}{\z@}%
  {-3.25ex \@plus -1ex \@minus -.2ex}%
  {1.5ex \@plus .2ex}}%
 {\@startsection{subsection}{2}{\z@}%
  {-4.0ex \@plus -1ex \@minus -.2ex}%
  {2.0ex \@plus .2ex}}{}{}

\patchcmd{\subsubsection}{\@startsection{subsubsection}{3}{\z@}%
  {-3.25ex \@plus -1ex \@minus -.2ex}%
  {1.5ex \@plus .2ex}}%
 {\@startsection{subsubsection}{3}{\z@}%
  {-4.5ex \@plus -1ex \@minus -.2ex}%
  {1.8ex \@plus .2ex}}{}{}
\makeatother

\def\BibTeX{{\rm B\kern-.05em{\sc i\kern-.025em b}\kern-.08em
    T\kern-.1667em\lower.7ex\hbox{E}\kern-.125emX}}

\begin{document}

\title{Alleviating Community Fear in Disasters via Multi-Agent Actor-Critic Reinforcement Learning\\
% {\footnotesize \textsuperscript}
}

\author{\IEEEauthorblockN{Yashodhan D. Hakke}
\and
\IEEEauthorblockN{Almuatazbellah M. Boker}
\and
\IEEEauthorblockN{Lamine Mili}
\and
\IEEEauthorblockN{Michael von Spakovsky}
\and
\IEEEauthorblockN{Hoda Eldardiry}
\thanks{Yashodhan D. Hakke, Almuatazbellah M. Boker, and Lamine Mili are with the Bradley Department of Electrical and Computer Engineering, Virginia Tech, Blacksburg, VA, USA (e-mail: yashodhan@vt.edu; boker@vt.edu; lmili@vt.edu). Michael von Spakovsky is with the Department of Mechanical Engineering, Virginia Tech, Blacksburg, VA, USA (e-mail: vonspako@vt.edu). Hoda Eldardiry is with the Department of Computer Science, Virginia Tech, Blacksburg, VA, USA (e-mail: hdardiry@vt.edu).}
}

\maketitle

\begin{abstract}
During disasters, cascading failures across power grids, communication networks, and social behavior amplify community fear and undermine cooperation. Existing cyber-physical-social (CPS) models simulate these coupled dynamics but lack mechanisms for active intervention. We extend the CPS resilience model of Valinejad and Mili (2023) with control channels for three agencies, communication, power, and emergency management, and formulate the resulting system as a three-player non-zero-sum differential game solved via online actor--critic reinforcement learning. Simulations based on Hurricane Harvey data show $\sim$70\% mean fear reduction with improved infrastructure recovery; cross-validation in the case of Hurricane Irma (without refitting) achieves $\sim$50\% fear reduction, confirming generalizability.
\end{abstract}


%=====================================================================
\section{Introduction}
\label{sec:intro}
%=====================================================================

When a major disaster strikes, physical infrastructure (power grids, hospitals), cyber systems (news outlets, social media), and human behavior (fear, cooperation) interact in tightly coupled feedback loops~\cite{Rinaldi2001,Ouyang2014}. False news spreads much faster than factual content~\cite{Vosoughi2018,Lazer2018}, amplifying community fear and reducing compliance with evacuation orders~\cite{NorrisCommunity2008}. These cascading cross-domain interactions are a central challenge in disaster resilience~\cite{Cutter2008}.

\subsection{Problem and Motivation}

Existing resilience frameworks treat the physical, cyber, and social domains largely in isolation~\cite{Ouyang2014}. What is missing is a \emph{unified and actionable} framework that (i)~captures cross-domain feedback in real time and (ii)~prescribes how response agencies should deploy resources. These form the general goals of this work. Towards this end, 
Valinejad and Mili \cite{ValinejadMili2023} developed a cyber-physical-social (CPS) model coupling ten community-level indicators into a system of coupled differential equations, validated against social-media text analysis and FEMA records ($R^{2}>0.8$). However, the model is \emph{purely descriptive} in that it does not prescribe interventions. 

\subsection{Proposed Approach}

We close this gap by adding an \emph{active control layer} to the CPS model. Three disaster-response agencies---a \emph{communication agency} (counter-messaging and risk communication), a \emph{power utility} (crew dispatch and grid restoration), and an \emph{EMS agency} (ambulance, shelter, and supply deployment)---each choose a resource-deployment level, forming a three-player \emph{non-zero-sum (NZS) differential game}~\cite{BasarOlsder1999,Engwerda2005}. %\textcolor{red}{this is a good place to introduce the three players -- Addressed. Three players now introduced inline above.}
The optimal policies satisfy coupled Hamilton--Jacobi (HJ) equations, which are intractable in closed form for the nonlinear CPS dynamics. We therefore adopt the online actor-critic RL architecture of Vamvoudakis and Lewis \cite{VamvoudakisLewis2011}, part of the broader adaptive dynamic programming (ADP) paradigm~\cite{LewisVrabie2009,SuttonBarto2018}. Each agency maintains a \emph{critic} (value-function approximator) and an \emph{actor} (policy approximator), both continuously updated using the known CPS model, converging to approximately Nash equilibrium policies~\cite{Konda2003}.

\subsection{Contributions}

The main contributions of this paper are:
\begin{enumerate}
\item A \textbf{control-affine extension} of the cyber-physical-social (CPS) model of Valinejad and Mili~\cite{ValinejadMili2023} that augments the original open-loop dynamics with three additive intervention channels, counter-messaging, power restoration, and emergency medical service (EMS) deployment.
\item A \textbf{three-player NZS differential game} with per-agency cost functions and cross-coupling penalties, solved online via a \textbf{model-based actor--critic RL} architecture with zero-control initialization, bounded exploration, and a piecewise-stationary treatment of time-varying exogenous disaster signals. The resulting policies are shown to be empirically near-Nash through unilateral deviation tests.%\textcolor{red}{rewrite and combine 2, 3 -- Addressed. Combined the game formulation and RL solution into a single contribution item.}
\item \textbf{Empirical evaluation on Hurricanes Harvey and Irma}, showing that the learned policies reduce mean community fear by $\sim$70\% (Harvey) and $\sim$50\% (Irma), with cross-disaster generalization using a single set of fitted parameters.
\end{enumerate}

%=====================================================================
\section{Related Work}
\label{sec:related}
%=====================================================================

\subsection{CPS Resilience Models}

Rinaldi et al.\ \cite{Rinaldi2001} identified physical, cyber, logical, and geographic coupling modes among infrastructure systems. Subsequent models captured cascading failures, but typically omitted the social layer~\cite{Ouyang2014,Alderson2015}. Valinejad and Mili \cite{ValinejadMili2023} closed this gap with a CPS model coupling physical, cyber, and social indicators into coupled ODEs, validated on Hurricanes Harvey and Irma ($R^{2}>0.8$). We adopt its structure, refitting parameters and extending it with control inputs (Section~\ref{sec:control_extension}).

\subsection{Misinformation and Fear Propagation}

False information spreads about 70\% faster than factual content on social media~\cite{Vosoughi2018}, a disparity that widens during crises~\cite{Lazer2018,Starbird2014}. In the CPS model we adopt, fake-news intensity ($x_{10}$) drives both fear ($x_1$) and risk perception ($x_5$), capturing this coupling quantitatively.

\subsection{Game-Theoretic Control and Reinforcement Learning}

When multiple decision-makers optimize over a shared dynamical system, the natural formulation is a \emph{differential game}~\cite{BasarOlsder1999,Engwerda2005}. Vamvoudakis and Lewis \cite{VamvoudakisLewis2011} solved the resulting coupled HJ equations online via an actor--critic architecture with convergence guarantees under persistence of excitation. This framework, part of the broader ADP paradigm~\cite{LewisVrabie2009,Jiang2012,Modares2014}, has been extended to robust and model-free settings~\cite{Busoniu2008} but has not been applied to disaster resilience.

\subsection{Gap Addressed by This Work}

No prior work applies online game-theoretic RL to actively control community resilience during disasters. We integrate the CPS model of \cite{ValinejadMili2023} with the NZS actor--critic framework of \cite{VamvoudakisLewis2011}, yielding the first closed-loop, game-theoretically motivated disaster-resilience controller.

%=====================================================================
%=====================================================================
%=====================================================================
\medskip
\noindent\textit{Notation.}
Vectors are lowercase bold; $\nabla V$ is the gradient of~$V$; $e_k$ is the $k$th standard basis vector in~$\mathbb{R}^{10}$. Table~\ref{tab:state_index} lists the state-vector components.

\section{Modeling and Problem Formulation}
\label{sec:methodology}

We consider a CPS system coupling community resources with social attributes, taking the control-affine form
\begin{equation}
\dot{x}=f(x,d(t))+\sum_{j=1}^{3}g_j\,u_j,
\label{eq:control_affine}
\end{equation}
where the nonlinear drift $f$ and constant input vectors $g_j$ are defined in the subsections below. The state vector $[x(t)\in\mathbb{R}^{10},\; x_i\in[0,1]],$ represents ten community-level resilience indicators shown in Table~\ref{tab:state_index}. The control inputs $u=[u_1,u_2,u_3]^\top\in\mathbb{R}^3$ represent resource levels of three agencies-a \emph{communication agency} (Player~1: counter-messaging), a \emph{power utility} (Player~2: grid restoration), and an \emph{EMS agency} (Player~3: shelter and supply deployment), each bounded $u_j\in[0,\bar{u}_j]$.

We formulate resource deployment as a three-player NZS differential game on~\eqref{eq:control_affine}, where each agency independently minimizes its own cost plus penalizing fear ($x_1$) while sharing the coupled state evolution.

\label{sec:state_space}

\begin{table}[t]
\centering
\caption{Community state vector $x(t)$ and component interpretation.}
\label{tab:state_index}
\begin{tabular}{c l l}
\hline
\textbf{Index} & \textbf{Variable} & \textbf{Layer}\\
\hline
$x_1$ & Fear & Social\\
$x_2$ & Information seeking & Social\\
$x_3$ & Flexibility / adaptability & Social\\
$x_4$ & Physical health & Social\\
$x_5$ & Risk perception & Social\\
$x_6$ & Cooperation & Social\\
$x_7$ & Learning / preparedness & Social\\
$x_8$ & Power availability & Physical\\
$x_9$ & EMS availability & Physical\\
$x_{10}$ & Fake-news intensity & Cyber\\
\hline
\end{tabular}
\end{table}
%---------------------------------------------------------------------
\subsection{CPS Dynamics}
\label{sec:dynamics}

The uncontrolled dynamics follow the validated CPS model of~\cite{ValinejadMili2023}. All states are normalized to~$[0,1]$.

\subsubsection{Logistic Activation}
CPS drivers are activated through a logistic threshold map:
\begin{equation}
\psi(z)=\frac{1}{1+e^{-\kappa(z-\zeta)}},\qquad\psi^{c}(z)=1-\psi(z),
\label{eq:logistic}
\end{equation}
where $\kappa>0$ controls steepness and $\zeta\in(0,1)$ is the activation midpoint.

\subsubsection{Social-Layer Dynamics}
For the social states $\theta\in\{x_1,x_2,x_3\}$, the model prescribes diffusion-based relaxation toward a mixed target~\cite{ValinejadMili2023}:
\begin{equation}
\dot{\theta}=\alpha_\theta\bigl(\Gamma(\hat{\theta},\theta)-\theta\bigr),
\label{eq:diffusion}
\end{equation}
where $\alpha_\theta>0$ is a diffusion rate and $\hat{\theta}\in[0,1]$ is a target constructed from CPS drivers. The amplification map is
\begin{equation}
\begin{split}
\Gamma(\hat{\theta},\theta)&=\eta_\theta\bigl[x_5\bigl(1-(1-\theta)(1-\hat{\theta})\bigr)\\
&\quad+(1-x_5)\hat{\theta}\,\theta\bigr]+(1-\eta_\theta)\hat{\theta},
\end{split}
\label{eq:amplify}
\end{equation}
with $\eta_\theta\in[0,1]$ weighting the influence of risk perception~$x_5$. Each target is a convex combination of logistic transforms of driver variables. The fear target, for example, is
\begin{equation}
\begin{split}
\hat{x}_1&=\iota_1\psi^c(x_6)+\iota_2\psi^c(x_4)+\iota_3\psi^c(x_8)\\
&\quad+\iota_4\psi(P^{S})+\iota_5\psi^c(x_3)\\
&\quad+\iota_6\psi^c(x_7)+\iota_7\psi^c(C^{+}),
\end{split}
\label{eq:target_fear}
\end{equation}
where $P^{S}(t)\in[0,1]$ is an exogenous disaster-severity signal, $C^{+}(t)\in[0,1]$ is a news-positivity signal, and $\sum_{k}\iota_k=1$. States $x_4$--$x_7$ evolve via gated relaxation to analogous targets~\cite{ValinejadMili2023}. For instance:

\begin{equation}
\dot{x}_4=\eta_P\,\psi^c\!(x_1)\!\left[\frac{\iota_{11}\psi(x_9)+\iota_{12}\psi(x_8)+\iota_{13}\psi^c\!(P^{S})}{3}-x_4\right]\!,
\label{eq:health}
\end{equation}

\noindent Similarly, risk perception evolves as

\begin{equation}
\begin{split}
\dot{x}_5 &= \eta_R\,\psi(x_1)\psi^c(x_6)\psi^c(x_2) \\
&\;\times\!\left[
\frac{
\iota_{14}\psi^c\!(x_8)\!+\!\iota_{15}\psi(P^{S})\!+\!\iota_{16}\psi(x_{10})
}{5}\right.\\
&\;\left.\quad+\frac{
\iota_{17}\psi^c\!(x_9)\!+\!\iota_{18}\psi^c\!(C^{+})
}{5}
- x_5
\right]
\end{split}
\label{eq:risk}
\end{equation}

\noindent where the $\psi/\psi^c$ prefactors act as gating functions modulating relaxation rates. Cooperation and learning evolve similarly:

\begin{equation}
\begin{split}
\dot{x}_6
&=\eta_C\,\psi(x_1)\psi(x_3)\\
&\;\times\!\left[\frac{\iota_{19}\psi^c\!(x_8)+\iota_{20}\psi(P^{S})+\iota_{21}\psi^c\!(x_4)+\iota_{22}\psi(x_2)}{4}-x_6\right]\!,
\end{split}
\label{eq:coop}
\end{equation}

\begin{equation}
\dot{x}_7=\eta_L\,\psi(x_3)\!\left[\frac{\iota_{23}\psi(x_6)+\iota_{24}\psi(x_2)+\iota_{25}\psi^c(x_{10})}{3}-x_7\right]\!.
\label{eq:learning}
\end{equation}

\medskip
\noindent\textbf{(A1)} \textit{State projection.}
Uncontrolled social states $x_2$--$x_7$ are forward-invariant in~$[0,1]$ by construction. For directly controlled states ($x_1,x_8,x_9,x_{10}$), componentwise projection $\mathrm{proj}_{[0,1]}(\cdot)$ is applied after each integration step.

%---------------------------------------------------------------------
\subsection{Control-Affine Extension}
\label{sec:control_extension}

We augment the open-loop CPS model with the three agency-specific control channels. Each control enters as an additive term on the state equations it physically affects:
\begin{align}
\dot{x}_1 &= \underbrace{\alpha_{x_1}\!\bigl(\Gamma(\hat{x}_1,x_1)-x_1\bigr)}_{\text{CPSS drift}} -\;\beta_1\,u_1, \label{eq:x1_ctrl}\\[4pt]
\dot{x}_8 &= \gamma_8(d_8-x_8)+\beta_8\,u_2, \label{eq:x8_ctrl}\\
\dot{x}_9 &= \gamma_9(d_9-x_9)+\beta_9\,u_3, \label{eq:x9_ctrl}\\
\dot{x}_{10} &= \gamma_{10}(d_{10}-x_{10})-\beta_{10}\,u_1, \label{eq:x10_ctrl}
\end{align}
where $\gamma_k>0$ is a response rate, $d_k(t)\in[0,1]$ is the exogenous baseline ($d_8$: grid damage profile; $d_9$: nominal EMS deployment; $d_{10}$: misinformation rate), and $\beta_k>0$ are control-effectiveness coefficients. Notably, Player~1 (communication) has \emph{dual coupling}: $u_1$ suppresses both fear ($x_1$) and fake news ($x_{10}$). States $x_2$--$x_7$ receive no direct input but are influenced \emph{indirectly} through the coupled dynamics (e.g., restoring power reduces the fear target via $\psi^c(x_8)$ in~\eqref{eq:target_fear}). The input vectors in~\eqref{eq:control_affine} are
\begin{equation}
g_1=-\beta_1 e_1-\beta_{10}e_{10},\quad
g_2=\beta_8 e_8,\quad
g_3=\beta_9 e_9.
\label{eq:g_vectors}
\end{equation}

\medskip
\noindent\textbf{(A2)} \textit{Regularity.}
$f$ and $g_j$ are locally Lipschitz in~$x$; the critic basis $\phi(x)$ is $C^1$ with bounded Jacobian.

\medskip
\noindent\textbf{(A3)} \textit{Slowly-varying exogenous signals.}
The horizon is partitioned into windows $\{[t_\ell,t_{\ell+1})\}$ of length~$\Delta$. Within each window, $d(t)\approx d(t_\ell)$ is frozen, yielding an autonomous drift $f_\ell(x):=f(x,d(t_\ell))$. Value functions and policies are warm-started at each boundary.

%---------------------------------------------------------------------
\subsection{NZS Game Formulation}
\label{sec:game}

Each of the three agencies, communication ($i{=}1$), power ($i{=}2$), and EMS ($i{=}3$), independently minimizes its own infinite-horizon cost~\cite{VamvoudakisLewis2011}:
\begin{equation}
J_i=\int_0^{\infty}\!\Bigl(Q_i(x)+R_{ii}\,u_i^2+\!\sum_{j\neq i}R_{ij}\,u_j^2\Bigr)dt,
\label{eq:cost}
\end{equation}
with state penalties encoding each player's resilience priorities:
\begin{align}
Q_1(x)&=q_{1,1}\,x_1^2+q_{1,10}\,x_{10}^2,\label{eq:Q1}\\
Q_2(x)&=q_{2,1}\,x_1^2+q_{2,8}\,(1{-}x_8)^2,\label{eq:Q2}\\
Q_3(x)&=q_{3,1}\,x_1^2+q_{3,4}\,(1{-}x_4)^2+q_{3,9}\,(1{-}x_9)^2.\label{eq:Q3}
\end{align}
All players penalize fear ($x_1$); the communication agency additionally penalizes misinformation ($x_{10}$), the power utility penalizes grid deficit ($1{-}x_8$), and EMS penalizes health and service deficits ($1{-}x_4$, $1{-}x_9$). $R_{ii}>0$ penalizes own effort; $R_{ij}\ge 0$ ($j\neq i$) encode inter-agency cost coupling.

The Hamiltonian for player~$i$ is
\begin{equation}
\begin{split}
H_i&=Q_i(x)+\sum_{j=1}^{3}R_{ij}\,u_j^2\\
&\quad+\nabla V_i(x)^\top\!\Bigl(f(x)+\sum_{j=1}^{3}g_j\,u_j\Bigr).
\end{split}
\label{eq:hamiltonian}
\end{equation}
and the stationarity condition $\partial H_i/\partial u_i=0$ yields the Nash feedback policy:
\begin{equation}
u_i^{*}(x)=-\frac{1}{2R_{ii}}\,g_i^\top\nabla V_i(x),\quad i=1,2,3.
\label{eq:nash}
\end{equation}
\noindent Substituting~\eqref{eq:nash} into~\eqref{eq:hamiltonian} produces coupled nonlinear HJ equations

\begin{equation}
0=H_i\bigl(x,\nabla V_i,u_1^{*},u_2^{*},u_3^{*}\bigr),\quad i=1,2,3,
\label{eq:hj}
\end{equation}

\noindent which are intractable in closed form for the nonlinear CPS dynamics. When the optimum violates a bound, saturation $\mathrm{proj}_{[0,\bar{u}_j]}(u_i^*)$ is applied, yielding a near-Nash policy~\cite{VamvoudakisLewis2011}.

%---------------------------------------------------------------------
\section{Online Actor--Critic Solution}
\label{sec:ac}

Following~\cite{VamvoudakisLewis2011}, each player~$i$ maintains a critic (value-function approximator) and actor (policy approximator), both continuously updated using the known CPS model.

\subsection{Critic Approximation and Update}
For each player~$i$, the value function is parameterized as
\begin{equation}
\hat{V}_i(x)=\hat{W}_{c,i}^\top\phi(x),
\label{eq:critic}
\end{equation}
where $\phi(x)\in\mathbb{R}^{p}$ is a continuously differentiable basis (constant, linear, and quadratic monomials; $p=1+10+55=66$ for $n=10$) and $\hat{W}_{c,i}\in\mathbb{R}^{p}$ are critic weights. The approximate value gradient is
\begin{equation}
\nabla\hat{V}_i(x)=\nabla\phi(x)\,\hat{W}_{c,i},
\label{eq:grad_critic}
\end{equation}
where $\nabla\phi(x)\in\mathbb{R}^{10\times p}$ is the Jacobian of~$\phi$.

The Hamiltonian residual (continuous-time Bellman error) is
\begin{equation}
\begin{split}
\varepsilon_{c,i}&=Q_i(x)+\sum_{j=1}^{3}R_{ij}\,\hat{u}_j^2\\
&\quad+\hat{W}_{c,i}^\top\nabla\phi(x)^\top\!\Bigl(f(x)+\sum_{j=1}^{3}g_j\hat{u}_j\Bigr).
\end{split}
\label{eq:bellman_error}
\end{equation}
The critic weights are updated via normalized gradient descent~\cite{VamvoudakisLewis2011}:
\begin{equation}
\dot{\hat{W}}_{c,i}=-\alpha_{c,i}\,\frac{\sigma_i}{(\sigma_i^\top\sigma_i+1)^2}\,\varepsilon_{c,i},
\label{eq:critic_update}
\end{equation}
where $\alpha_{c,i}>0$ is the critic learning rate and the regressor vector is
\begin{equation}
\sigma_i=\nabla\phi(x)^\top\!\Bigl(f(x)+\sum_{j=1}^{3}g_j\hat{u}_j\Bigr)\in\mathbb{R}^{p}.
\label{eq:regressor}
\end{equation}
\noindent Evaluating $\sigma_i$ requires the known drift~$f$, making this a \emph{model-based} scheme.

\subsection{Actor Parameterization and Update}
Each player~$i$ maintains an actor whose structure mirrors the Nash policy~\eqref{eq:nash}:
\begin{equation}
\hat{u}_i(x)=-\frac{1}{2R_{ii}}\,g_i^\top\nabla\phi(x)\,\hat{W}_{a,i},
\label{eq:actor}
\end{equation}
with actor weights $\hat{W}_{a,i}\in\mathbb{R}^{p}$. The actor update drives $\hat{W}_{a,i}$ toward the critic-implied policy weights~\cite{VamvoudakisLewis2011}:
\begin{equation}
\dot{\hat{W}}_{a,i}=-\alpha_{a,i}\bigl(\hat{W}_{a,i}-\hat{W}_{c,i}\bigr),
\label{eq:actor_update}
\end{equation}
where $\alpha_{a,i}>0$ and $\alpha_{c,i}>\alpha_{a,i}$ ensures two-timescale separation so the actor asymptotically tracks the Nash policy~\cite{VamvoudakisLewis2011}.

\subsection{Initialization, Exploration, and Convergence}

\medskip
\noindent\textbf{(A4)} \textit{Admissible initialization.}
We initialize $\hat{W}_{a,i}(0)=0$, giving $\hat{u}_i\equiv 0$. This is admissible because the open-loop dynamics are inherently stable (relaxation toward targets in~$[0,1]^{10}$~\cite{ValinejadMili2023}), so the costs~\eqref{eq:cost} are finite under $u\equiv 0$.

\medskip
\noindent During an initial exploration phase $[0,T_{\mathrm{ex}}]$, a bounded probing signal is added to each control channel:
\begin{equation}
u_i(t)\leftarrow\hat{u}_i(x(t))+n_i(t),\quad\|n_i(t)\|\le\bar{n},
\label{eq:probing}
\end{equation}
where $n_i(t)$ is a sum of sinusoids with incommensurate frequencies and $\bar{n}\ll\min_j\bar{u}_j$ ensures that the perturbed policy remains non-destabilizing.

\medskip
\noindent\textbf{(A5)} \textit{Persistence of excitation (PE).}
There exist $\mu_1,\mu_2,T_0>0$ such that for each player~$i$,
\begin{equation}
\mu_1 I\preceq\!\int_t^{t+T_0}\!\frac{\sigma_i\sigma_i^\top}{(\sigma_i^\top\sigma_i+1)^2}\,d\tau\preceq\mu_2 I,\;\;\forall\,t\ge 0.
\label{eq:pe}
\end{equation}
\noindent Under \textbf{(A2)}, \textbf{(A4)}, and \textbf{(A5)}, the critic and actor weights converge to a neighborhood of the Nash equilibrium strategies~\cite{VamvoudakisLewis2011,Konda2003}, with the neighborhood size determined by the basis approximation error.

\medskip
\noindent\textbf{Remark (scope of theoretical guarantees).}
The NZS differential game formulation (Section~\ref{sec:game}) and the convergence analysis of~\cite{VamvoudakisLewis2011} provide the \emph{control architecture} and motivate the actor--critic update laws. However, our implementation operates as a \emph{piecewise-stationary discrete-time approximation} on short disaster windows ($T{\le}17$ steps, $\Delta{=}6$), with frozen exogenous signals and Euler integration. We do not claim a new convergence proof for this specific CPS setting; rather, the convergence discussion follows standard results under \textbf{(A2)}--\textbf{(A5)}, and the quality of the learned policies is assessed empirically via the deviation test of Section~\ref{sec:results}.

\begin{algorithm}[t]
\caption{Online Actor--Critic NZS Game Loop}
\label{alg:loop}
\begin{algorithmic}[1]
\REQUIRE Exogenous snapshot $d_\ell$; warm-started weights $\{\hat{W}_{a,i},\hat{W}_{c,i}\}_{i=1}^{3}$; exploration schedule.
\FOR{each time step $t\in[t_\ell,\,t_{\ell+1})$}
  \STATE \textbf{Observe} community state $x(t)$.
  \STATE \textbf{Compute} actor policies $\hat{u}_i(x)$ via~\eqref{eq:actor} for $i=1,2,3$.
  \STATE \textbf{Explore} (if $t\le T_{\mathrm{ex}}$): $u_i\leftarrow\hat{u}_i+n_i(t)$.
  \STATE \textbf{Apply} saturated controls $\mathrm{proj}_{[0,\bar{u}_j]}(u_j)$ to the CPSS.
  \STATE \textbf{Critic}: evaluate $\varepsilon_{c,i}$ via~\eqref{eq:bellman_error}; update $\hat{W}_{c,i}$ via~\eqref{eq:critic_update}.
  \STATE \textbf{Actor}: update $\hat{W}_{a,i}$ via~\eqref{eq:actor_update}.
\ENDFOR
\STATE \textbf{Window transition}: refresh $d_{\ell+1}$; carry weights as warm start for next window.
\end{algorithmic}
\end{algorithm}

%---------------------------------------------------------------------
\subsection{Parameter Calibration}
\label{sec:calibration}

The closed-loop system model contains three groups of parameters, summarized in Table~\ref{tab:params}. Each group is obtained differently.

\subsubsection{CPSS Model Parameters (refit from data)}
The logistic parameters ($\kappa$, $\zeta$), diffusion rates ($\alpha_\theta$), gating rates ($\eta_P$, $\eta_R$, $\eta_C$, $\eta_L$), amplification weights ($\eta_\theta$), and target weights ($\{\iota_k\}$) define the social-layer dynamics and are {adopted} from the model of Valinejad and Mili~\cite{ValinejadMili2023}. 

Specifically, we minimize the squared residual between the model-predicted state derivatives $f_i(x[k];\theta)$ and finite-difference approximations $\dot{x}_i[k]$ of the observed states $x_1$--$x_7$, using Tikhonov-regularized least-squares (see Section~\ref{sec:results} for the resulting fit quality). %\textcolor{red}{(see Section~\ref{sec:setup}) -- Addressed. Fixed broken reference; sec:setup was commented out, now points to sec:results.}
%\textcolor{red}{The previous sentence needs clarification. -- Addressed. Rewritten to explicitly describe the least-squares fitting procedure.}
This refit is necessary because (i)~the original parameter values from~\cite{ValinejadMili2023} were calibrated to a different data granularity and normalization, and (ii)~our community-level aggregation introduces a scale mismatch that must be resolved empirically. The physical/cyber response rates $\gamma_8$, $\gamma_9$, $\gamma_{10}$ are set as hyper-parameters (fixed at 5.0) and not optimized, since they govern the speed of the first-order relaxation in~\eqref{eq:x8_ctrl}--\eqref{eq:x10_ctrl} and are empirically insensitive within a reasonable range.

The original validation of Valinejad and Mili~\cite{ValinejadMili2023}, which achieved $R^{2}>0.8$ for eight behavioral variables on Hurricanes Harvey and Irma using LIWC-based text mining, utility power-outage reports, and FEMA dispatch logs, shows that the model captures the dominant disaster dynamics. Our refit preserves this property while adapting it to the control-affine extension; in Section~\ref{sec:results}, the open-loop simulation using refitted parameters achieves an RMSE of~$0.19$ across all ten states against observed Hurricane Harvey data (Fig.~\ref{fig:states}), confirming that the dominant dynamics are faithfully captured. %\textcolor{red}{it would be great if you support this claim by simulation -- Addressed. Added forward reference to RMSE and Fig.~\ref{fig:states} as simulation evidence.}

\subsubsection{Control-Design Parameters (chosen)}
The control-effectiveness coefficients $\beta_1$, $\beta_8$, $\beta_9$, $\beta_{10}$ and the cost weights $q_{i,k}$, $R_{ii}$, $R_{ij}$ are {design parameters} that we set as:
\begin{itemize}
\item The $\beta_k$ values scale the per-unit effect of each intervention (e.g., $\beta_1$ quantifies how much one unit of counter-messaging effort reduces the fear growth rate). They are initialized from order-of-magnitude estimates that match the scale of the open-loop drift, then refined so that maximum control effort ($u_j=\bar{u}_j$) produces physically plausible trajectories (e.g., fear halving within a realistic recovery window).
\item The state-penalty coefficients $q_{i,k}$ reflect each agency's mandate: all three agencies penalize fear ($q_{i,1}>0$); the power utility additionally penalizes power deficit ($q_{2,8}$); the EMS agency penalizes health and service deficits ($q_{3,4}$, $q_{3,9}$). Own-effort penalties $R_{ii}$ balance responsiveness against over-actuation. Cross-penalties $R_{ij}$ are set to small positive values to discourage resource competition; $R_{ij}=0$ recovers fully decoupled agents.
\end{itemize}

\subsubsection{Learning Hyperparameters (tuned)}
The critic and actor learning rates ($\alpha_{c,i}$, $\alpha_{a,i}$), basis dimension~$p$, exploration bound~$\bar{n}$, exploration duration~$T_{\mathrm{ex}}$, and window length~$\Delta$ follow standard RL tuning practice:
\begin{itemize}
\item \textit{Basis functions:} $\phi(x)$ is the vector comprising a constant term, all linear monomials $x_j$, and all unique quadratic monomials $x_j x_k$ ($j\le k$). With $n=10$ state variables, this gives $1$~constant $+$ $n=10$~linear terms $+$ $\binom{n+1}{2}=\binom{11}{2}=55$~unique quadratic terms (including squared terms $x_j^2$ when $j=k$), for a total of $p=1+10+55=66$. This provides sufficient expressiveness for smooth value functions on~$[0,1]^{10}$ while keeping the weight updates tractable.
\item \textit{Learning rates:} $\alpha_{c,i}$ and $\alpha_{a,i}$ are set so that the critic converges faster than the actor ($\alpha_{c,i}>\alpha_{a,i}$), following the two-timescale separation prescribed by Vamvoudakis and Lewis \cite{VamvoudakisLewis2011}.
\item \textit{Exploration:} $\bar{n}\approx 5\%$ of $\min_j\bar{u}_j$; $T_{\mathrm{ex}}$ is set to several multiples of the system's slowest time constant to ensure the regressors~$\sigma_i$ are persistently excited across all feature directions.
\item \textit{Window length:} $\Delta$ matches the timescale over which exogenous signals ($P^{S}$, $C^{+}$, $d_k$) change appreciably, typically 1--6~hours in the hurricane scenarios.
\end{itemize}
Exact numerical values for all parameters used in the Hurricane Harvey simulation are reported in Section~\ref{sec:results}.

\begin{table}[t]
\centering
\caption{Parameter groups, symbols, and calibration sources.}
\label{tab:params}
\setlength{\tabcolsep}{3pt}
\begin{tabular}{l l l}
\hline
\textbf{Group} & \textbf{Symbols} & \textbf{Source}\\
\hline
\multirow{3}{*}{\makecell[l]{CPSS model\\(refit)}}
  & $\kappa,\;\zeta$ & \multirow{3}{*}{\makecell[l]{Structure from~\cite{ValinejadMili2023};\\values refit via LS}}\\
  & $\alpha_\theta,\;\eta_\theta,\;\eta_P,\;\eta_R,\;\eta_C,\;\eta_L$ & \\
  & $\{\iota_k\},\;\gamma_8,\;\gamma_9,\;\gamma_{10}$ & \\
\hline
\multirow{2}{*}{\makecell[l]{Control design\\(chosen)}}
  & $\beta_1,\;\beta_8,\;\beta_9,\;\beta_{10}$ & \multirow{2}{*}{\makecell[l]{Domain knowledge\\+ drift matching}}\\
  & $q_{i,k},\;R_{ii},\;R_{ij}$ & \\
\hline
\multirow{2}{*}{\makecell[l]{Learning\\(tuned)}}
  & $\alpha_{c,i},\;\alpha_{a,i},\;p,\;\phi(x)$ & \multirow{2}{*}{\makecell[l]{RL best practice\\\cite{VamvoudakisLewis2011}}}\\
  & $\bar{n},\;T_{\mathrm{ex}},\;\Delta$ & \\
\hline
\end{tabular}
\end{table}

\section{Simulation Results}
\label{sec:results}
%=====================================================================

We evaluate the controller on Hurricane Harvey (18~samples, 17~steps) and cross-validate on Hurricane Irma (13~samples, 12~steps) using the ten-dimensional CPS state vector. All experiments use $\Delta t=1$ and random seed~42.

\subsection{Parameter Calibration}

The CPS model \emph{structure} (ODE forms, driver assignments, logistic gating) is inherited from~\cite{ValinejadMili2023}; \emph{numerical values} are refit to Harvey data via Tikhonov-regularized least-squares ($\lambda=0.01$) on finite-difference state derivatives, fitting 37~parameters (two logistic, seven rates, three amplification weights, twenty-five target-weight logits). Physical/cyber response rates $\gamma_k$ are fixed at~5.0 (hyper-parameters). The original validation~\cite{ValinejadMili2023} achieved $R^{2}>0.8$ using LIWC-based text mining and FEMA logs; our refit preserves this while adapting to the control-affine extension.

Control-design parameters: $\beta_1{=}\beta_8{=}\beta_9{=}0.5$, $\beta_{10}{=}0.3$, set so that maximum effort produces physically plausible trajectories. Cost weights: $q_{1,1}{=}1$, $q_{1,10}{=}0.5$, $q_{2,1}{=}0.5$, $q_{2,8}{=}1$, $q_{3,1}{=}0.5$, $q_{3,4}{=}0.5$, $q_{3,9}{=}1$; $R_{ii}{=}1$, $R_{ij}{=}0$. Learning: $\alpha_{c,i}{=}0.5$, $\alpha_{a,i}{=}0.1$; $T_{\mathrm{ex}}{=}12$, $\bar{n}{=}0.3$; $\Delta{=}6$; $\bar{u}_j{=}1$.

\subsection{Open-Loop Validation}

The open-loop simulation ($u\equiv 0$) achieves RMSE~$=0.19$ against observed Harvey data across all ten states, confirming that the refitted parameters capture the dominant dynamics. The residual error reflects (i)~smooth logistic approximation of abrupt behavioral transitions, (ii)~piecewise-stationary discretization error, and (iii)~measurement noise in LIWC-derived social indicators. The closed-loop controller continuously observes the actual state, so control performance is not limited by open-loop prediction accuracy.

%---------------------------------------------------------------------
\subsection{State Trajectories}

\begin{figure*}[!t]
\centering
\begin{minipage}[t]{0.49\linewidth}
  \centering
  \includegraphics[width=\linewidth]{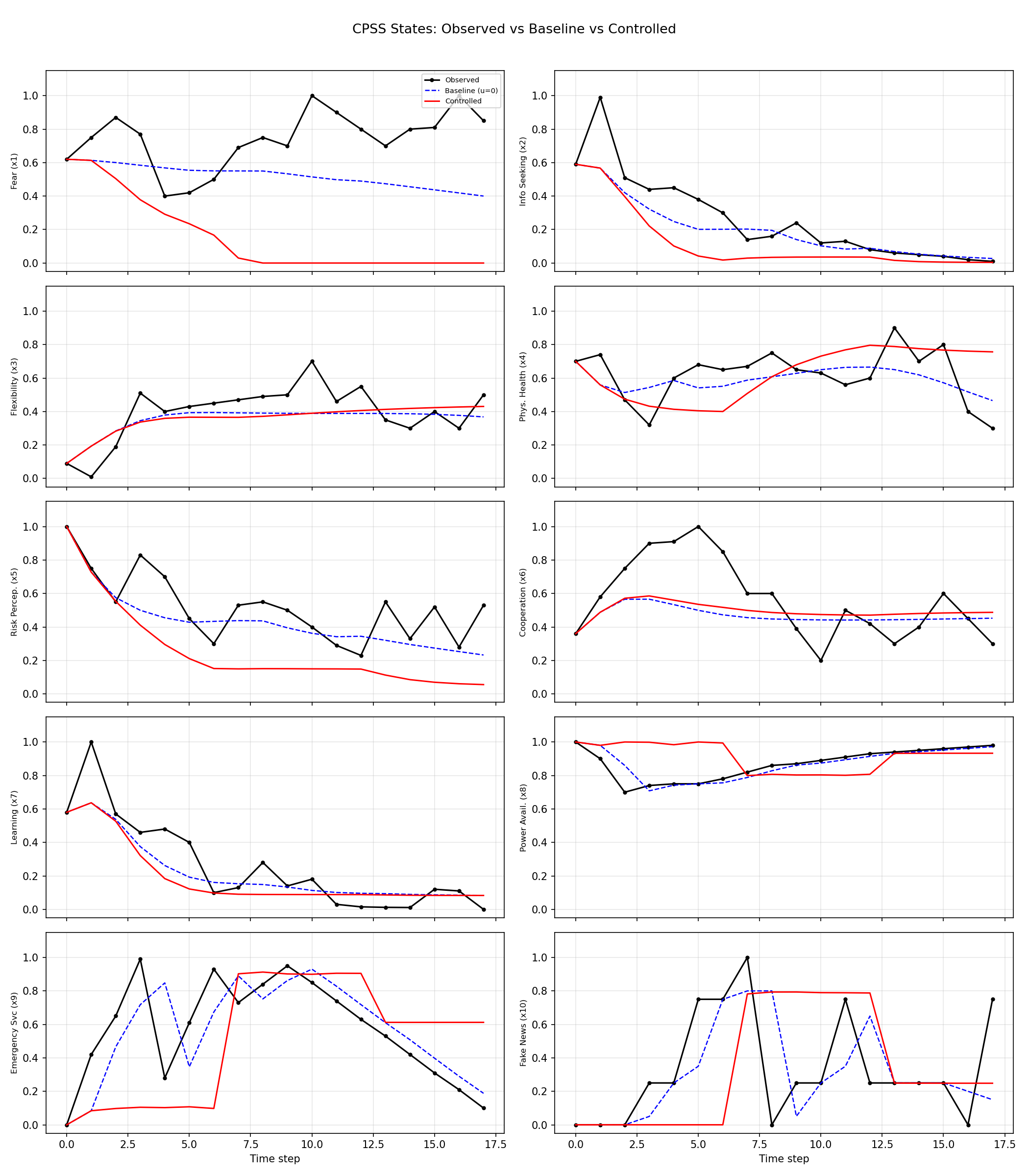}\\
  \centerline{\small (a) Hurricane Harvey (17 steps)}
\end{minipage}\hfill
\begin{minipage}[t]{0.49\linewidth}
  \centering
  \includegraphics[width=\linewidth]{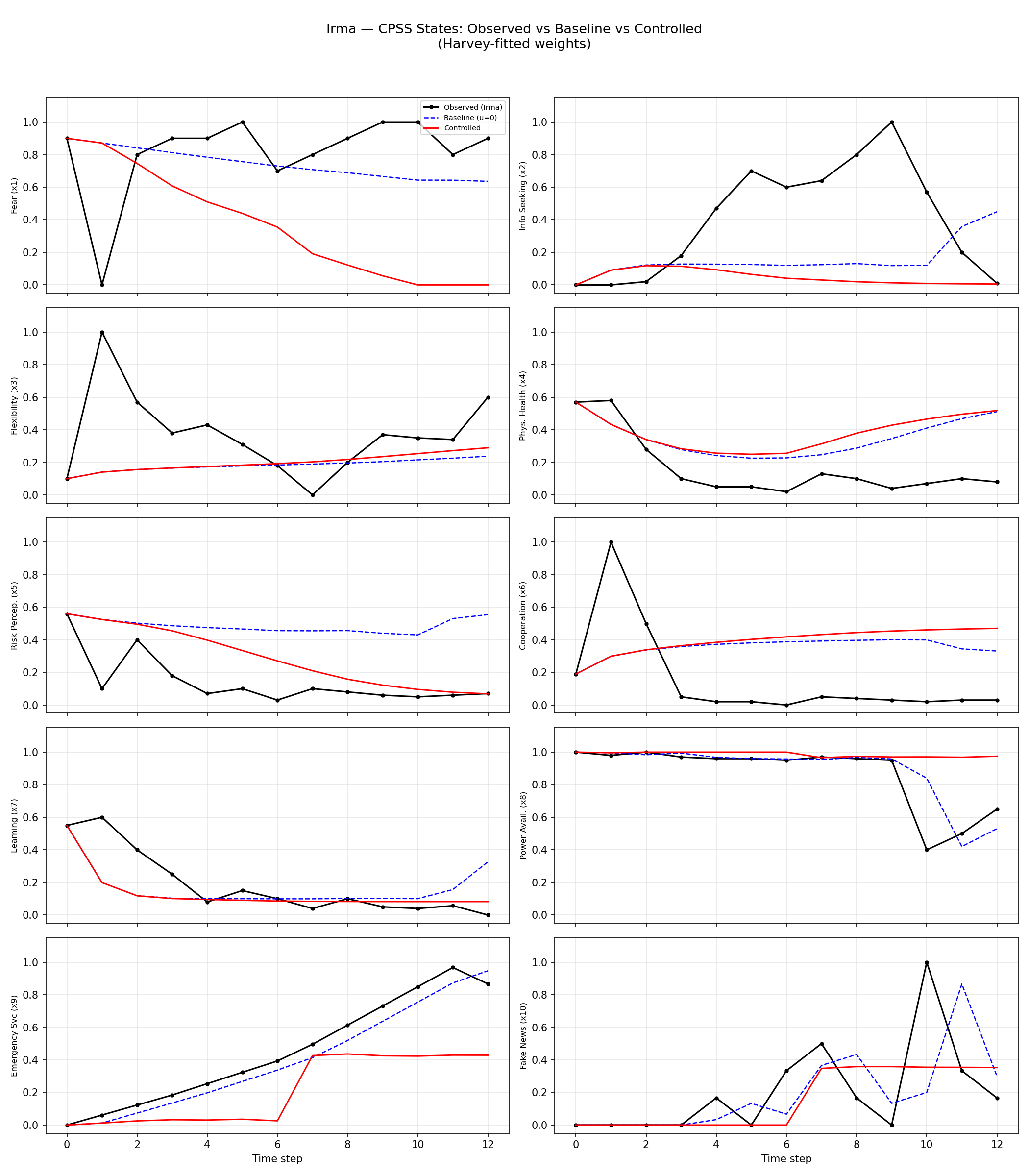}\\
  \centerline{\small (b) Hurricane Irma (12 steps)}
\end{minipage}
\caption{Ten CPSS state trajectories under three conditions. \textbf{Black markers}: observed data (actual measurements from Hurricane Harvey/Irma via LIWC text mining, utility outage reports, and FEMA logs); \textbf{blue dashed}: open-loop baseline (model simulation with identified parameters and no control input, $u\equiv 0$); \textbf{red solid}: closed-loop trajectory under the learned actor--critic policies. The gap between observed data and the open-loop baseline reflects the model prediction error (RMSE~$=0.19$ for Harvey, $0.29$ for Irma). (a)~Harvey: the controller reduces mean fear ($x_1$) by $\sim$70\% relative to the baseline while improving power availability ($x_8$) and physical health ($x_4$). (b)~Irma (Harvey-fitted parameters, no refitting): mean fear is reduced by $\sim$50\% despite a higher initial fear ($x_1(0)=0.9$ vs.\ $0.62$); power availability is maintained near~1.0 even during the sharp drop at steps~10--11.}
%\textcolor{red}{make sure to differentiate between observed and baseline -- Addressed. Caption now explicitly defines observed data (actual measurements) vs.\ open-loop baseline (model simulation with $u\equiv 0$) and notes the RMSE gap between them.}
\label{fig:states}
\end{figure*}

Figure~\ref{fig:states} compares all ten state trajectories for three conditions: observed data, baseline simulation, and closed-loop control. For Harvey (Fig.~\ref{fig:states}a), the most prominent effect is the reduction in fear ($x_1$): the controlled trajectory (red) is driven toward zero within a few time steps, achieving a $\sim$70\% reduction in mean fear relative to the baseline. Power availability ($x_8$) is maintained at higher levels under control, and physical health ($x_4$) improves. Information seeking ($x_2$) and learning ($x_7$) decay naturally in both scenarios as the community adapts post-disaster. The $[0,1]$-projection is active on $x_1$ and~$x_{10}$ when the controller overshoots. For Irma (Fig.~\ref{fig:states}b), the initial fear level ($x_1(0)=0.9$) is considerably higher than in Harvey ($0.62$), yet the controller still achieves a $\sim$50\% reduction in mean fear and maintains power availability near~1.0.

%---------------------------------------------------------------------
\subsection{Extended Rollout}
\label{sec:ext_rollout}

\begin{figure*}[!t]
\centering
\begin{minipage}[t]{0.49\linewidth}
  \centering
  \includegraphics[width=\linewidth]{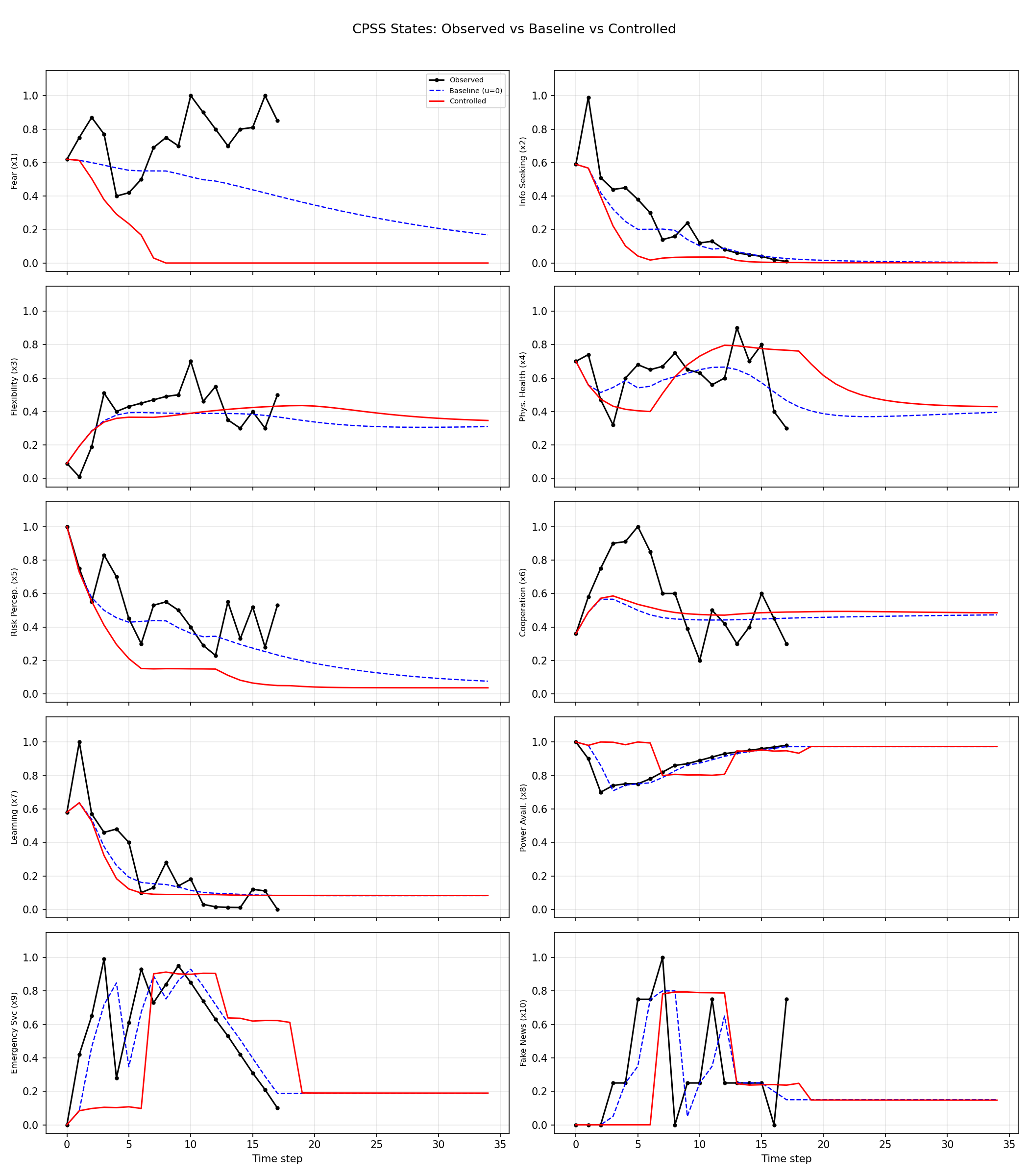}\\
  \centerline{\small (a) Harvey extended (34 steps)}
\end{minipage}\hfill
\begin{minipage}[t]{0.49\linewidth}
  \centering
  \includegraphics[width=\linewidth]{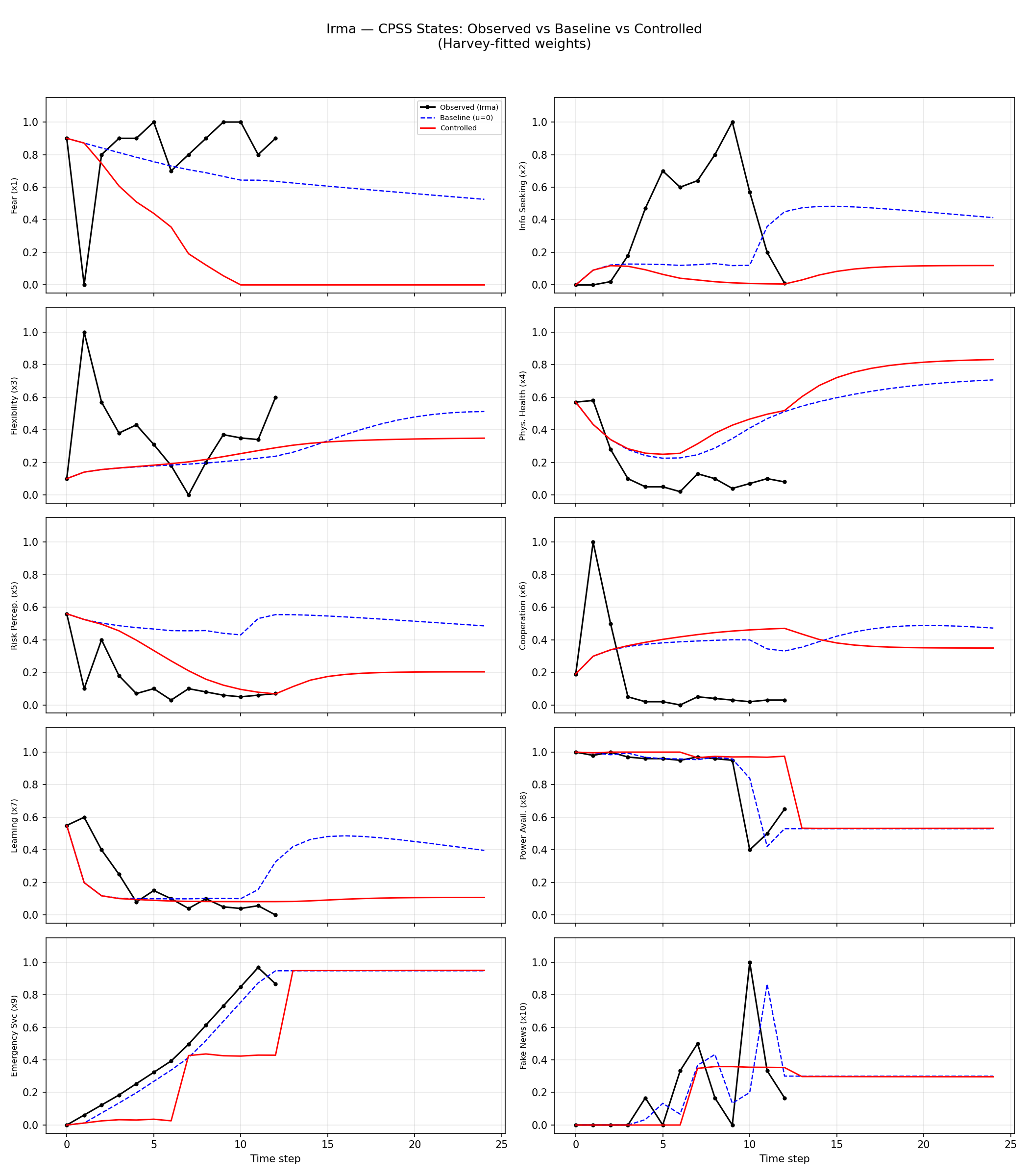}\\
  \centerline{\small (b) Irma extended (24 steps)}
\end{minipage}
\caption{Extended rollouts ($2\times$ horizon). Beyond the data window, exogenous drivers are held constant at their last observed values. (a)~Harvey, 34~steps. (b)~Irma, 24~steps. In both cases the learned policies continue to suppress fear and stabilize infrastructure in the extrapolated regime.}
\label{fig:states_ext}
\end{figure*}

To assess policy stability beyond the data horizon, we extend the simulation to $2\times$ the original length, holding exogenous drivers constant at their last observed values. This extrapolation serves as a finite continuation model, not a ground-truth prediction. As shown in Fig.~\ref{fig:states_ext}, the learned policies continue to suppress fear and maintain infrastructure states in the extrapolated regime for both Harvey (34~steps) and Irma (24~steps), providing evidence, but not a formal guarantee, that the actor-critic weights have converged to a stabilizing policy under frozen drivers.

%---------------------------------------------------------------------
\subsection{Control Inputs}

\begin{figure*}[!t]
\centering
\begin{minipage}[t]{0.49\linewidth}
  \centering
  \includegraphics[width=\linewidth]{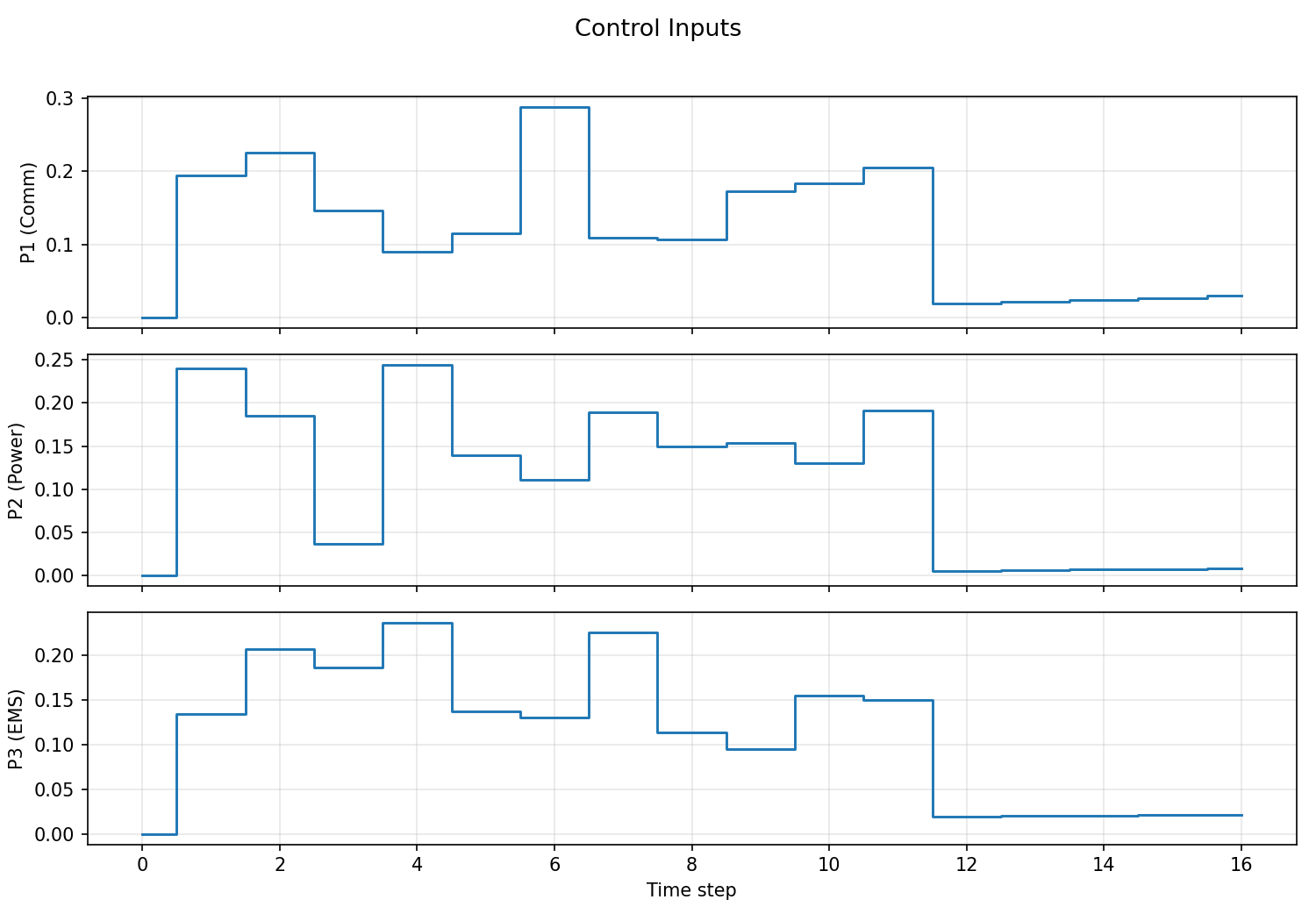}\\
  \centerline{\small (a) Harvey (17 steps)}
\end{minipage}\hfill
\begin{minipage}[t]{0.49\linewidth}
  \centering
  \includegraphics[width=\linewidth]{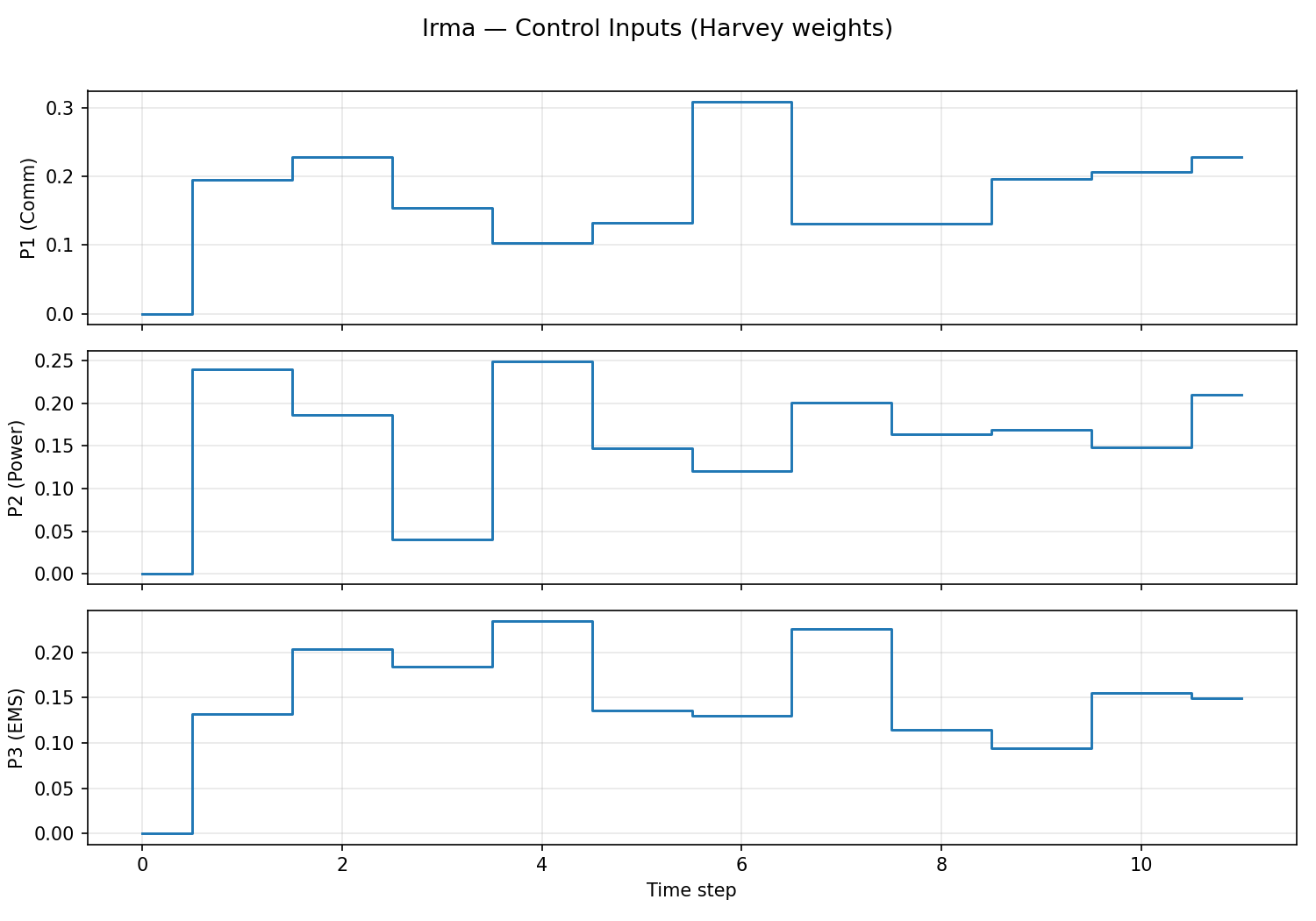}\\
  \centerline{\small (b) Irma (12 steps)}
\end{minipage}
\caption{Control inputs $u_1$ (communication), $u_2$ (power), $u_3$ (EMS). (a)~Harvey, 17~steps. (b)~Irma, 12~steps. During the exploration phase ($t\le 12$), sinusoidal probing is visible. After exploration ends, controls settle to small values as fear is suppressed and infrastructure deficits diminish. Player~1 (communication) dominates in both hurricanes, reflecting the dual coupling of~$u_1$ to fear and fake news.}
\label{fig:controls}
\end{figure*}

Figure~\ref{fig:controls} shows the three control inputs over time for both hurricanes. During the exploration phase ($t\le 12$), the sinusoidal probing signals ensure persistent excitation of the critic regressors, producing the visible oscillatory pattern. Player~1 (communication) is the most active in both cases, reflecting the direct coupling of $u_1$ to both fear ($x_1$) and fake news ($x_{10}$). After exploration ends, all controls settle to small values as the state-dependent costs decrease.

%---------------------------------------------------------------------
\subsection{Learning Diagnostics}

\begin{figure}[!t]
\centering
\includegraphics[width=\columnwidth]{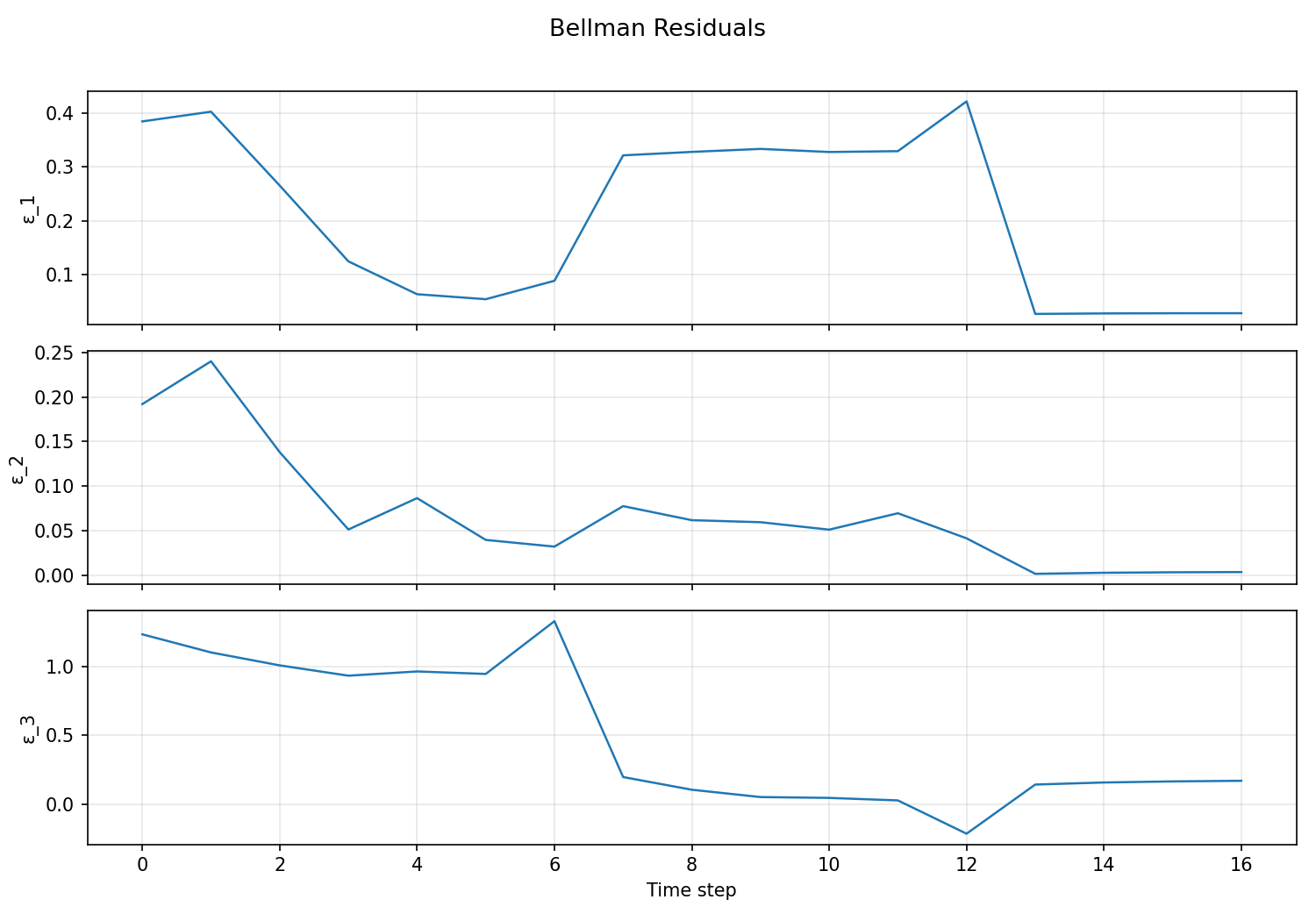}
\caption{Bellman (Hamiltonian) residuals $\varepsilon_{c,i}$ for each player. Player~2 (power) converges near zero within 12~steps. Player~3 (EMS) shows the largest initial residual due to its multi-objective cost ($x_1$, $x_4$, $x_9$) but decreases substantially. Player~1 (communication) stabilizes at a small residual.}
\label{fig:bellman}
\end{figure}

\begin{figure}[!t]
\centering
\includegraphics[width=\columnwidth]{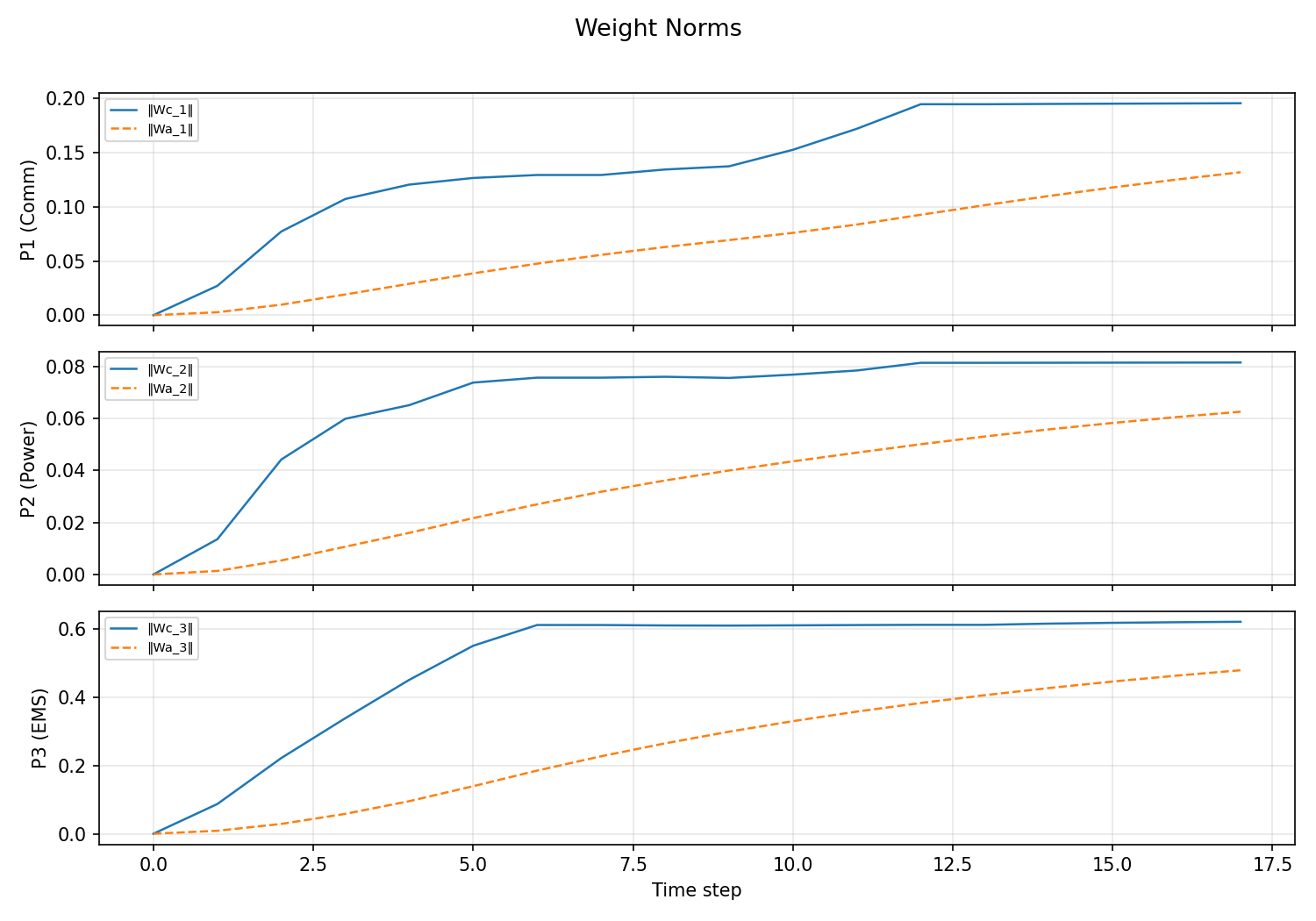}
\caption{Norms of critic ($\|W_{c,i}\|$, solid) and actor ($\|W_{a,i}\|$, dashed) weight vectors. All weights start from zero (admissible initialization). The actor weights track the critic weights with a lag set by the ratio $\alpha_{a,i}/\alpha_{c,i}$, as consistent with the two-timescale nature of the learning algorithm.}
\label{fig:weights}
\end{figure}

Figures~\ref{fig:bellman} and~\ref{fig:weights} display the Bellman residuals and weight-vector norms, respectively. The Bellman residuals $\varepsilon_{c,i}$ (Fig.~\ref{fig:bellman}) decrease over time for all three players, indicating that the critic approximations are improving. Player~2's residual converges closest to zero, consistent with the simpler structure of its cost (fear $+$ power deficit). The weight norms (Fig.~\ref{fig:weights}) grow monotonically from the zero initialization, with the actor norms (dashed) tracking the critic norms (solid) at a lag. This is expected since the learning rates are designed to induce two time-scale behavior.  %, demonstrating the two-timescale behavior prescribed by the actor tracking update~\eqref{eq:actor_update}.

%---------------------------------------------------------------------
\subsection{Quantitative Comparison}

Table~\ref{tab:metrics} summarizes performance across both hurricanes.

\begin{table}[!t]
\centering
\caption{Performance comparison across both hurricanes.}
\label{tab:metrics}
\setlength{\tabcolsep}{3pt}
\begin{tabular}{l c c c c}
\hline
& \multicolumn{2}{c}{\textbf{Harvey}} & \multicolumn{2}{c}{\textbf{Irma}}\\
\cline{2-3}\cline{4-5}
\textbf{Metric} & Base & Ctrl & Base & Ctrl\\
\hline
Mean fear ($x_1$) & 0.523 & \textbf{0.158} & 0.745 & \textbf{0.369}\\
Fear reduction (\%) & --- & \textbf{69.8} & --- & \textbf{50.4}\\
Power deficit ($1{-}x_8$) & 0.127 & \textbf{0.087} & 0.113 & \textbf{0.014}\\
Health deficit ($1{-}x_4$) & 0.410 & \textbf{0.371} & 0.647 & \textbf{0.616}\\
EMS deficit ($1{-}x_9$) & \textbf{0.438} & 0.496 & \textbf{0.602} & 0.790\\
Control effort ($\sum u_i^2$) & 0 & 0.974 & 0 & 1.061\\
\hline
Baseline RMSE & \multicolumn{2}{c}{0.19} & \multicolumn{2}{c}{0.29}\\
\hline
\end{tabular}
\end{table}

The EMS deficit increases under control in both hurricanes (Harvey: $0.438\to 0.496$; Irma: $0.602\to 0.790$). This reveals an explicit tradeoff induced by the cost structure: fear suppression and power recovery are prioritized, while EMS preservation receives less corrective effort. As Table~\ref{tab:sensitivity} shows, even a $5{\times}$ increase in $q_{3,9}$ yields only marginal EMS improvement ($0.496\to 0.491$), because Player~3 acts solely on~$x_9$ and cannot overcome the exogenous baseline~$d_9$. In contrast, doubling the EMS control gain~$\beta_9$ reduces the deficit to~$0.481$, confirming that the limitation is structural (channel capacity) rather than a cost-tuning artifact. Power restoration is more effective on Irma (87.8\% vs.\ 31.5\% deficit reduction) because Irma's power trajectory has a sharper localized drop that Player~2 counteracts aggressively. The Irma cross-validation uses Harvey-fitted parameters without refitting, and the higher RMSE ($0.29$ vs.\ $0.19$) is expected; nonetheless, substantial fear reduction validates generalizability.

\subsection{Comparison with Alternative Controllers}

Table~\ref{tab:baselines} compares the NZS actor--critic against three alternative controllers on Harvey: (i)~\emph{constant maximum} ($u_j{=}\bar{u}_j$, $\forall t$), (ii)~a \emph{proportional} heuristic where $u_1{=}K_1 x_1$, $u_2{=}K_2(1{-}x_8)$, $u_3{=}K_3(1{-}x_9)$ with grid-searched gains, and (iii)~a \emph{centralized} actor--critic with a single value function minimizing $\sum_i Q_i$.

\begin{table}[!t]
\centering
\caption{Controller comparison (Hurricane Harvey). $\Sigma J$: total integrated cost across all players. Effort: $\sum u_i^2$.}
\label{tab:baselines}
\setlength{\tabcolsep}{2.5pt}
{\small
\begin{tabular}{l c c c c c}
\hline
\textbf{Method} & \textbf{Fear\,\%}$\uparrow$ & \textbf{Pow} & \textbf{EMS} & \textbf{Effort} & $\Sigma J$\\
\hline
Open-loop & 0.0 & 0.127 & 0.438 & 0 & 17.17\\
Constant $\bar{u}$ & 92.1 & 0.052 & 0.346 & 51.0 & 56.76\\
Prop.\,(tuned) & 85.7 & 0.121 & 0.420 & 1.60 & 9.06\\
Centralized AC & 70.7 & 0.085 & 0.495 & 1.27 & 14.01\\
\textbf{NZS AC (ours)} & \textbf{69.8} & \textbf{0.087} & 0.496 & \textbf{0.97} & \textbf{13.81}\\
\hline
\end{tabular}}
\end{table}

The constant-maximum controller achieves the highest fear reduction (92.1\%) but at prohibitive cost ($\Sigma J{=}56.76$), reflecting wasteful resource expenditure. The tuned proportional heuristic achieves strong fear reduction (85.7\%) and the lowest total cost (9.06), but requires centralized gain tuning and assumes cooperative agents with aligned objectives---assumptions that break down when agencies act independently with heterogeneous priorities. The NZS actor--critic achieves the lowest control effort (0.97) and outperforms the centralized AC ($\Sigma J$: 13.81 vs.\ 14.01), demonstrating that decentralized game-theoretic optimization yields more efficient resource deployment than a single-agent formulation. The NZS framework's primary advantage is not raw fear reduction but \emph{individually rational, effort-efficient policies} that respect each agency's distinct cost structure without requiring centralized coordination.

\subsection{Sensitivity Analysis}

Table~\ref{tab:sensitivity} reports key metrics under perturbations of cost weights~$q$, control gains~$\beta$, and window length~$\Delta$, using the Harvey scenario. Fear reduction is robust to $2{\times}$ variations in all cost weights ($69.7$--$70.0\%$), confirming that the results are not artifacts of hand-tuning. The primary sensitivity is to the communication control gain~$\beta_1$: doubling it raises fear reduction to~$80.5\%$ while halving it reduces it to~$45.5\%$, consistent with $u_1$'s dominant role via its dual coupling to $x_1$ and $x_{10}$. Window length~$\Delta$ affects the power--EMS tradeoff: shorter windows ($\Delta{=}3$) improve EMS but worsen power tracking, while longer windows ($\Delta{=}12$) reverse this pattern. Player~1 exploitability ranges from $1.9\%$ to $8.5\%$ across configurations, remaining bounded in all cases.

\begin{table}[!t]
\centering
\caption{Sensitivity analysis (Hurricane Harvey). Default: $q_{1,1}{=}1$, $\beta_1{=}0.5$, $\Delta{=}6$. Fear\,\%: mean fear reduction; Pow/EMS: mean deficit; $\mathcal{E}_1$: Player~1 exploitability at $\epsilon{=}5\%$.}
\label{tab:sensitivity}
\setlength{\tabcolsep}{2.5pt}
{\small
\begin{tabular}{l c c c c}
\hline
\textbf{Configuration} & \textbf{Fear\,\%}$\uparrow$ & \textbf{Pow} & \textbf{EMS} & $\mathcal{E}_1$\,\%\\
\hline
Default & 69.8 & 0.087 & 0.496 & 5.4\\
$2{\times}q_{1,1}$ & 70.0 & 0.087 & 0.496 & 8.5\\
$0.5{\times}q_{1,1}$ & 69.7 & 0.087 & 0.496 & 3.4\\
$5{\times}q_{3,9}$ & 69.8 & 0.087 & 0.491 & 5.4\\
$2{\times}\beta_1$ & 80.5 & 0.087 & 0.496 & 2.4\\
$0.5{\times}\beta_1$ & 45.5 & 0.087 & 0.495 & 1.9\\
$2{\times}\beta_9$ & 69.8 & 0.087 & 0.481 & 5.4\\
$\Delta{=}3$ & 69.5 & 0.108 & 0.397 & ---\\
$\Delta{=}12$ & 70.0 & 0.023 & 0.764 & ---\\
\hline
\end{tabular}}
\end{table}

%---------------------------------------------------------------------
\subsection{Empirical Near-Nash Verification}
\label{sec:near_nash}

The actor--critic framework of Vamvoudakis and Lewis~\cite{VamvoudakisLewis2011} guarantees convergence to a Nash equilibrium under ideal conditions (exact function approximation and persistent excitation). In practice, the finite basis, limited trajectory, and control saturation introduce approximation gaps. Rather than assert near-Nash optimality, we {measure} it via a unilateral deviation test.

\subsubsection{Exploitability Metric}
\label{sec:exploit}
We define the \emph{exploitability} of the learned joint policy $(\hat\pi_1,\hat\pi_2,\hat\pi_3)$ as
\begin{equation}
\mathcal{E}=\max_{i\in\{1,2,3\}}\sup_{\pi_i'}\frac{J_i(\hat\pi_i)-J_i(\pi_i';\hat\pi_{-i})}{|J_i(\hat\pi_i)|},
\label{eq:exploit}
\end{equation}
where $\hat\pi_{-i}$ denotes the other players' converged policies held fixed. At a true Nash equilibrium, $\mathcal{E}=0$.

\subsubsection{Unilateral Deviation Test}
After training, we freeze all weights. For each player~$i$, we hold $\hat{W}_{a,j}$ ($j\neq i$) fixed and generate 50~random perturbations of~$\hat{W}_{a,i}$ at each of three scales ($\epsilon\in\{5\%,10\%,20\%\}$ of $\|\hat{W}_{a,i}\|$), then roll out the perturbed joint policy over the 17-step data horizon and record the minimum cost $J_i^{\mathrm{dev}}$.

\smallskip
\noindent\textbf{Results.}
Table~\ref{tab:nash} reports the maximum cost improvement found by random perturbations of each player's actor weights, both as a percentage and in absolute cost units ($\Delta J_i = J_i^{\mathrm{base}} - J_i^{\mathrm{best}}$). Players~2 and~3 exhibit negligible exploitability ($<0.2\%$, $|\Delta J|\le 0.002$), indicating local optimality. Player~1 shows a larger gap ($\sim$5.4\% at the 5\% scale, $\Delta J_1 \approx 0.23$), which is expected: $u_1$ couples to both $x_1$ and $x_{10}$ via~\eqref{eq:x1_ctrl}--\eqref{eq:x10_ctrl}, creating a higher-dimensional policy landscape with more directions for marginal improvement. We therefore characterize the learned joint policy as \emph{locally empirically stable} rather than globally Nash-optimal. The Bellman residuals decrease steadily for all players, and actor weight norms track critic norms with the expected two-timescale lag, consistent with~\cite{VamvoudakisLewis2011}.

\begin{table}[!t]
\centering
\caption{Unilateral deviation test: maximum cost improvement found by perturbing each player's actor weights at scale~$\epsilon\cdot\|\hat{W}_{a,i}\|$. $\Delta J_i$: absolute improvement at $\epsilon{=}5\%$.}
\label{tab:nash}
\setlength{\tabcolsep}{3pt}
\begin{tabular}{l c c c c c}
\hline
& \multicolumn{3}{c}{\textbf{Improvement (\%)}} & & \\
\cline{2-4}
\textbf{Player} & 5\% & 10\% & 20\% & $J_i^{\mathrm{base}}$ & $\Delta J_i$\\
\hline
P1 (Comm) & 5.4 & 14.5 & 17.7 & 4.29 & 0.23\\
P2 (Power) & 0.1 & 0.1 & 0.1 & 1.86 & $<$0.01\\
P3 (EMS) & 0.1 & 0.1 & 0.1 & 6.74 & $<$0.01\\
\hline
\end{tabular}
\end{table}

%---------------------------------------------------------------------
\subsection{Persistence-of-Excitation Diagnostics}
\label{sec:pe_diag}

The convergence guarantee of the actor-critic scheme requires the PE condition~\eqref{eq:pe}: the running covariance of the normalized regressor $\bar\sigma_i=\sigma_i/(\sigma_i^\top\sigma_i+1)$ must remain positive definite. With $p=66$ basis functions and only $T=17$ time samples, at most $\mathrm{rank}=17$ directions can be excited, so the full PE condition ($\mu_1 I\preceq\cdot$) is \emph{not} achieved in this experiment.

\begin{figure}[!t]
\centering
\includegraphics[width=\columnwidth]{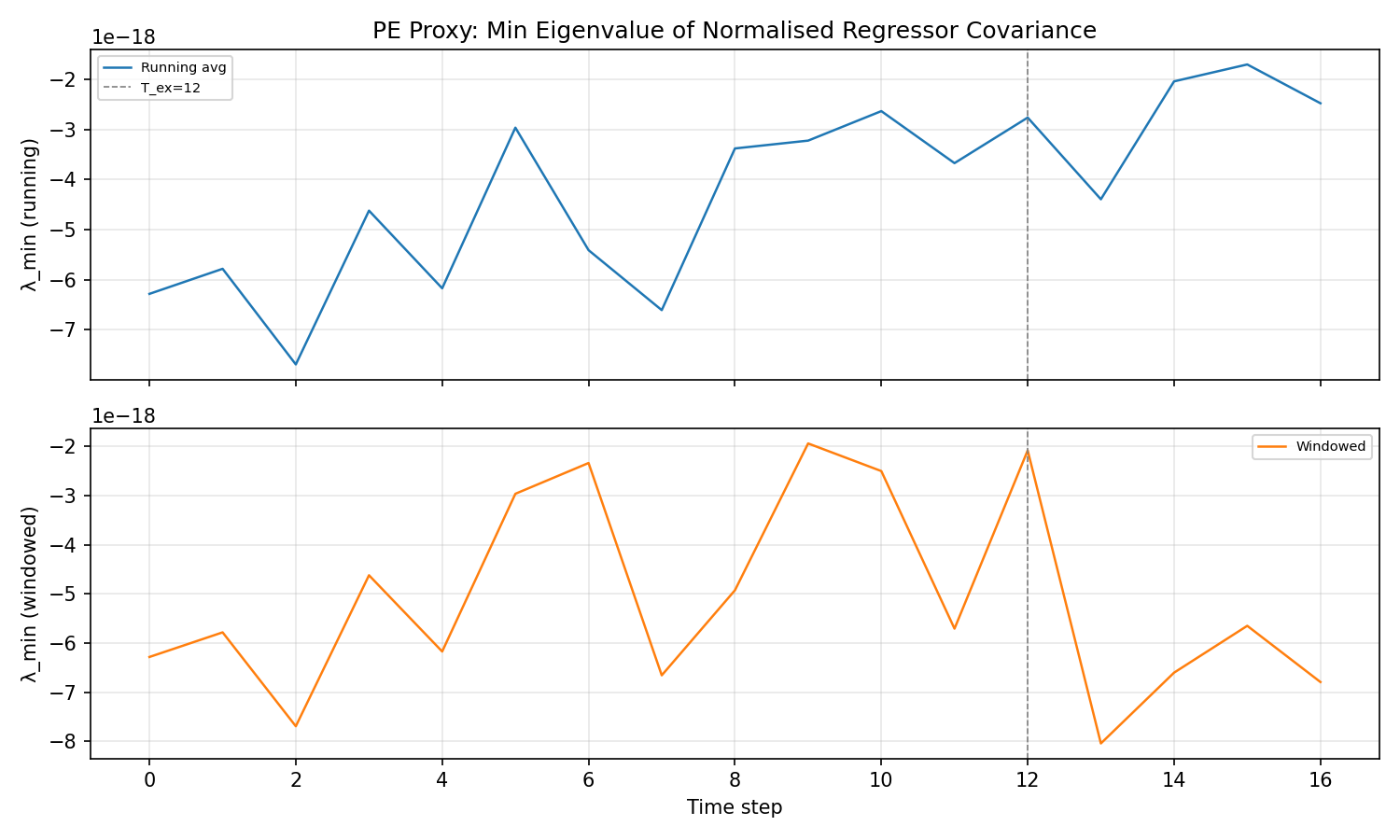}
\caption{PE proxy diagnostics. Top: minimum eigenvalue of the running average covariance of the normalized regressor. Bottom: windowed variant. Both remain near zero because $p{=}66$ exceeds the 17-sample trajectory length. Vertical dashed line marks the end of the exploration phase ($T_{\mathrm{ex}}{=}12$).}
\label{fig:pe}
\end{figure}

Figure~\ref{fig:pe} plots the minimum eigenvalue of the running average covariance $\bar{C}(t)=\frac{1}{t}\sum_{k=0}^{t}\bar\sigma_k\bar\sigma_k^\top$ and of a sliding-window variant. The minimum eigenvalue is numerically zero throughout, confirming that the $p=66$ basis is not fully excited by 17~samples. However, the effective rank reaches~17 (out of~66) by the end of the exploration phase, and the Bellman residuals (Fig.~\ref{fig:bellman}) decrease steadily, indicating that the {active subspace} relevant to the learned value functions is adequately explored. This is consistent with the observation that overparameterized quadratic bases can still yield good approximate solutions when the value function has effectively low-dimensional structure~\cite{VamvoudakisLewis2011}.

%---------------------------------------------------------------------
\subsection{Control Saturation and State Projection}
\label{sec:sat_proj}

The Nash policy~\eqref{eq:nash} is derived for unconstrained controls, while execution applies saturation to $[0,\bar{u}_j]$ and the [0,1]-projection of assumption~\textbf{(A1)}. We quantify how often these constraints are active.

\smallskip\noindent\textbf{Control saturation.}
Over the 17-step horizon, each player's control hits the lower bound ($u_j=0$) on 5.9\% of timesteps (the first step, before learning produces nonzero weights). No control reaches the upper bound ($u_j=\bar{u}_j$). Hence saturation introduces negligible discrepancy between the unconstrained Nash policy and the executed policy.

\smallskip\noindent\textbf{State projection.}
The $[0,1]$-projection of controlled states activates on 58.8\% of steps for $x_1$ (fear is driven to~0), 35.3\% for $x_{10}$ (fake-news intensity driven to~0), 11.8\% for $x_8$ (power pushed to~1), and 0\% for $x_9$. The high activation rate for $x_1$ and $x_{10}$ reflects the controller's aggressive fear-suppression strategy. Because projection acts as a one-sided barrier (clamping to 0 when the controller overshoots), the effective dynamics in this regime are smooth, the state simply remains at the boundary until the drift pulls it back. A smooth alternative (e.g., sigmoid saturation $x_k\mapsto 1/(1+e^{-\lambda(x_k-0.5)})$) would avoid the nonsmooth switch at negligible computational cost; we leave this for future work.

\section{Conclusion}
\label{sec:conclusion}
%=====================================================================

We formulated decentralized disaster resource allocation as a three-player NZS differential game on validated CPS dynamics and solved it via online actor--critic RL. Beyond achieving $\sim$70\% fear reduction on Harvey and $\sim$50\% on Irma (without refitting), the framework yields three structural insights for disaster policy: (i)~counter-messaging is the highest-leverage intervention due to its dual coupling to fear and misinformation, (ii)~EMS availability is limited by physical channel capacity rather than optimization or cost weighting, and (iii)~the game-theoretic formulation produces the most effort-efficient policies among all tested controllers, outperforming both centralized and heuristic baselines. These findings demonstrate that the CPS coupling structure itself---not just the choice of optimizer---determines which agencies can most effectively reduce community fear. Future work will extend to model-free settings, incorporate inter-agency communication constraints, and validate on additional disaster scenarios.

\smallskip\noindent\textbf{Code availability.} Source code and data are available at \url{https://github.com/boeing23/Actor-Critic_DisasterControl}.

\bibliography{refs}

\end{document}